\documentclass[10pt,twocolumn,letterpaper]{article}
\usepackage{cvpr}
\usepackage{times}
\usepackage{epsfig}
\usepackage{graphicx}
\usepackage{amsmath}
\usepackage{amssymb}
\usepackage{pbox}    
\usepackage{epstopdf}
\usepackage{subfigure}
\usepackage{xspace}
\usepackage{comment}
\usepackage{lipsum}   
\usepackage{jeffe}
\usepackage{booktabs}
\usepackage{multirow}
\usepackage[percent]{overpic}

\newcommand{\shape}{S}
\newcommand{\image}{I}
\newcommand{\network}{\mathbb{G}}
\newcommand{\prob}{\mathcal{P}}

\newcommand{\para}[1]{\noindent{\bf #1}}

\newcommand{\denselist}{\itemsep 0pt\parsep=0pt\partopsep 0pt\vspace{-2pt}}
\newcommand{\bitem}{\begin{itemize}\denselist}
\newcommand{\eitem}{\end{itemize}}
\newcommand{\benum}{\begin{enumerate}\denselist}
\newcommand{\eenum}{\end{enumerate}}
\newcommand{\bdescr}{\begin{description}\denselist}
\newcommand{\edescr}{\end{description}}

\cvprfinalcopy 

\def\cvprPaperID{190} 

\ifcvprfinal\fi
\begin{document}

\title{A Point Set Generation Network for \\ 3D Object Reconstruction from a Single Image}

\author{Haoqiang Fan \thanks{equal contribution} \\
Institute for Interdisciplinary\\ Information Sciences\\
Tsinghua University\\
{\tt\small fanhqme@gmail.com}
\and
Hao Su${}^*$~~~~~~~~~~~~~~~Leonidas Guibas\\
Computer Science Department\\
\\
Stanford University\\
{\tt\small \{haosu,guibas\}@cs.stanford.edu}
}

\maketitle

\begin{abstract}
  Generation of 3D data by deep neural network has been attracting increasing attention in the research community. The majority of extant works resort to regular representations such as volumetric grids or collection of images; however, these representations obscure the natural invariance of 3D shapes under geometric transformations, and also suffer from a number of other issues.
In this paper we address the problem of 3D reconstruction from a single image, generating a straight-forward form of output -- point cloud coordinates. Along with this problem arises a unique and interesting issue, that the groundtruth shape for an input image may be ambiguous. Driven by this unorthodox output form and the inherent ambiguity in groundtruth, we design architecture, loss function and learning paradigm that are novel and effective. Our final solution is a conditional shape sampler, capable of predicting multiple plausible 3D point clouds from an input image.
In experiments not only can our system outperform state-of-the-art methods on single image based 3d reconstruction benchmarks; but it also shows strong performance for 3d shape completion and promising ability in making multiple plausible predictions.
\end{abstract}

\section{Introduction}
\label{sec:intro}


As we try to duplicate the successes of current deep convolutional architectures in the 3D domain, we face a fundamental representational issue. Extant deep net architectures for both discriminative and generative learning in the signal domain are well suited to data that is regularly sampled, such as images, audio, or video. However, most common 3D geometry representations, such as 2D meshes or point clouds are not regular structures and do not easily fit into architectures that exploit such regularity for weight sharing, etc.  That is why the majority of extant works on using deep nets for 3D data resort to either volumetric grids or collections of images (2D views of the geometry). Such representations, however, lead to difficult trade offs between sampling resolution and net efficiency. Furthermore, they enshrine quantization artifacts that obscure natural invariances of the data under rigid motions, etc.
\begin{figure}[t!]
	\centering
	\includegraphics[width=\linewidth]{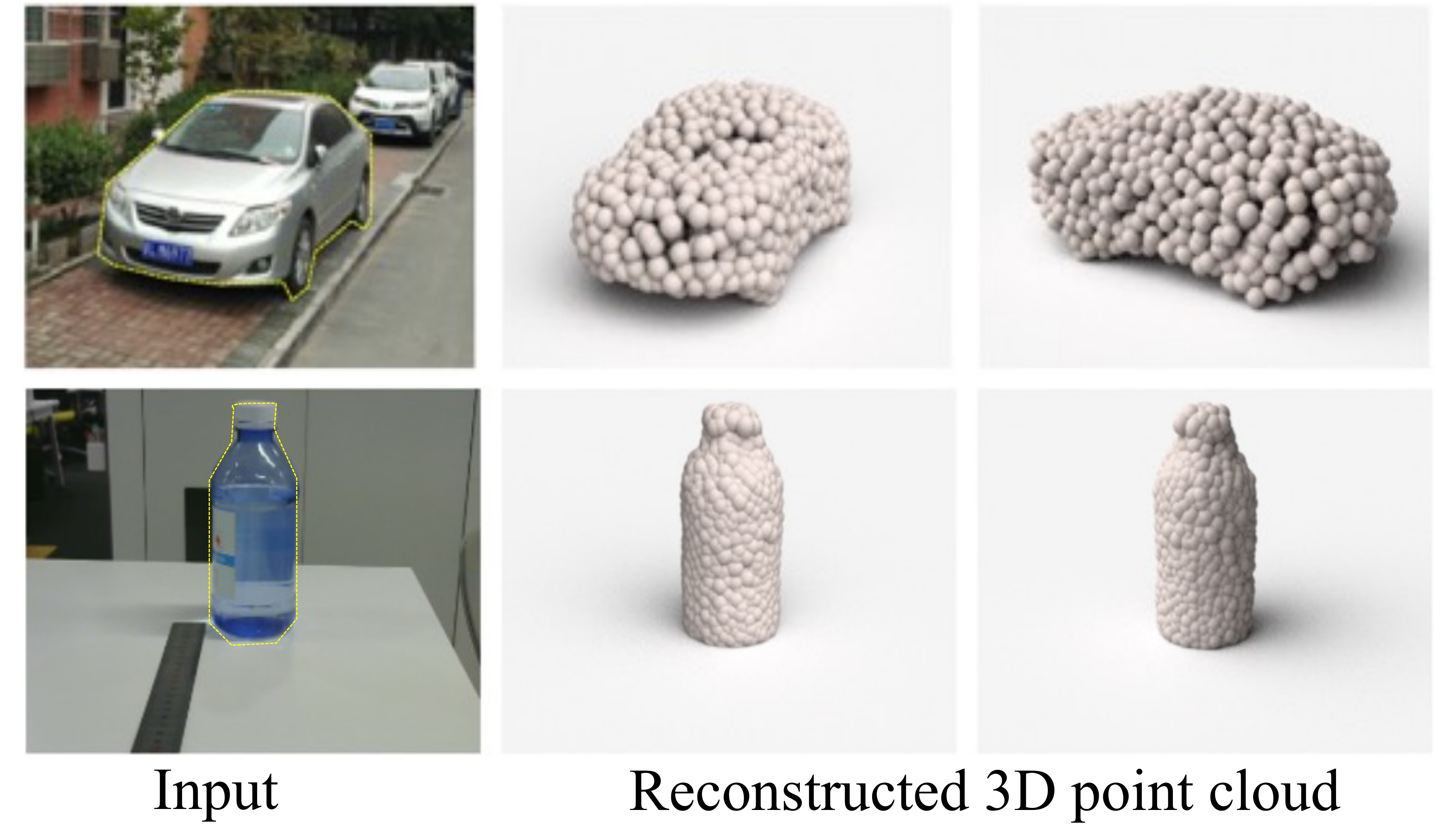}
	\caption{A 3D point cloud of the {\bf complete} object can be reconstructed from a single image. Each point is visualized as a small sphere. The reconstruction is viewed at two viewpoints ($0^{\circ}$ and $90^{\circ}$ along azimuth). A segmentation mask is used to indicate the scope of the object in the image.}
	\label{fig:teaser}
	\vspace{-1em}
\end{figure}

In this paper we address the problem of generating the 3D geometry of an object based on a single image of that object. We explore generative networks for 3D geometry based on a point cloud representation. A point cloud representation may not be as efficient in representing the underlying continuous 3D geometry as compared to a CAD model using geometric primitives or even a simple mesh, but for our purposes it has many advantages, A point cloud is a simple, uniform structure that is easier to learn, as it does not have to encode multiple primitives or combinatorial connectivity patterns. In addition, a point cloud allows simple manipulation when it comes to geometric transformations and deformations, as connectivity does not have to be updated. Our pipeline infers the point positions in a 3D frame determined by the input image and the inferred viewpoint position.

Given this unorthodox network output, one of our challenges is how to measure loss during training, as the same geometry may admit different point cloud representations at the same degree of approximation. Unlike the usual $L_2$ type losses, we use the solution of a transportation problem based on the Earth Mover's distance (EMD), effectively solving an assignment problem. We exploit an approximation to the EMD to provide speed as well as ensure differentiability for end-to-end training.

Our approach effectively attempts to solve the ill-posed problem of 3D structure recovery from a single projection using certain learned priors. The network has to estimate depth for the visible parts of the image and hallucinate the rest of the object geometry, assessing the plausibility of several different completions. From a statistical perspective, it would be ideal if we can fully characterize the landscape of the ground truth space, or be able to sample plausible candidates accordingly. If we view this as a regression problem, then it has a rather unique and interesting feature arising from inherent object ambiguities in certain views. These are situations where there are multiple, equally good 3D reconstructions of a 2D image, making our problem very different from classical regression/classification settings, where each training sample has a unique ground truth annotation. In such settings the proper loss definition can be crucial in getting the most meaningful result.

Our final algorithm is a conditional sampler, which samples plausible 3D point clouds from the estimated ground truth space given an input image. Experiments on both synthetic and real world data verify the effectiveness of our method. 
Our contributions can be summarized as follows:
\bitem
  \item We are the first to study the point set generation problem by deep learning; 
  \item On the task of 3D reconstruction from a single image, we apply our point set generation network and significantly outperform state of the art;
  \item We systematically explore issues in the architecture and loss function design for point generation network;
  \item We propose a principled formulation and solution to address the groundtruth ambiguity issue for the 3D reconstruction from single image task.
\eitem

\section{Related Work}
\label{sec:related}


\paragraph{3D reconstruction from single images}
While most researches focus on multi-view geometry such as SFM and SLAM~\cite{haming2010structure, fuentes2015visual}, ideally, one expect that 3D can be reconstructed from the abundant single-view images. 

Under this setting, however, the problem is ill-posed and priors must be incorporated. Early work such as ShapeFromX~\cite{horn1989obtaining,aloimonos1988shape} made strong assumptions over the shape or the environment lighting conditions. \cite{hoiem2005automatic,saxena2009make3d} pioneered the use of learning-based approach for simple geometric structures. Coarse correspondences in an image collection can also be used for rough 3D shape estimation~\cite{shapesKarTCM15,carreira2016lifting}. As commodity 3D sensors become popular, RGBD database has been built and used to train learning-based systems~\cite{eigen2014depth,Fouhey13}. Though great progress has been made, these methods still cannot robustly reconstruct complete and quality shapes from single images. Stronger shape priors are missing.

Recently, large-scale repositories of 3D CAD models, such as ShapeNet~\cite{shapenet2015}, have been introduced. They have great potential for 3D reconstruction tasks. For example, \cite{su2014estimating,huang2015single} proposed to deform and reassemble existing shapes into a new model to fit the observed image. These systems rely on high-quality image-shape correspondence, which is a challenging and ill-posed problem itself.

More relevant to our work is \cite{choy20163d}. 
Given a single image, they use a neural network to predict the underlying 3D object as a 3D volume. There are two key differences between our work and \cite{choy20163d}: First, the predicted object in \cite{choy20163d} is a 3D volume; whilst ours is a point cloud. As demonstrated and analyzed in Sec~\ref{sec:exp:rgb}, point set forms a nicer shape space for neural networks, thus the predicted shapes tend to be more complete and natural. Second, we allow multiple reconstruction candidates for a single input image. This design reflects the fact that a single image cannot fully determine the reconstruction of a 3D shape. 

\paragraph{Deep learning for geometric object synthesis} In general, the field of how to predict geometries in an end-to-end fashion is quite a virgin land. In particular, our output, 3D point set, is still not a typical object in the deep learning community. A point set contains orderless samples from a metric-measure space. Therefore, equivalent classes are defined up to a permutation; in addition, the ground distance must be taken into consideration. To our knowledge, we are not aware of prior deep learning systems with the abilities to predict such objects. 


\section{Problem and Notations}
\begin{figure*}[t!]
  \centering
  \begin{overpic}[width=\linewidth,unit=1mm]{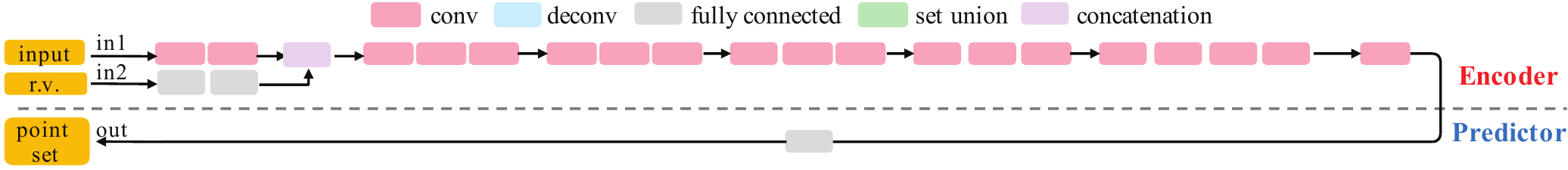} 
    \put(47, -0.5){vanilla version}
  \end{overpic}  
  \qquad\\  
  \qquad\\
  \begin{overpic}[width=\linewidth, trim={0cm, 0cm, 0cm, 0.47cm}, clip,unit=1mm]{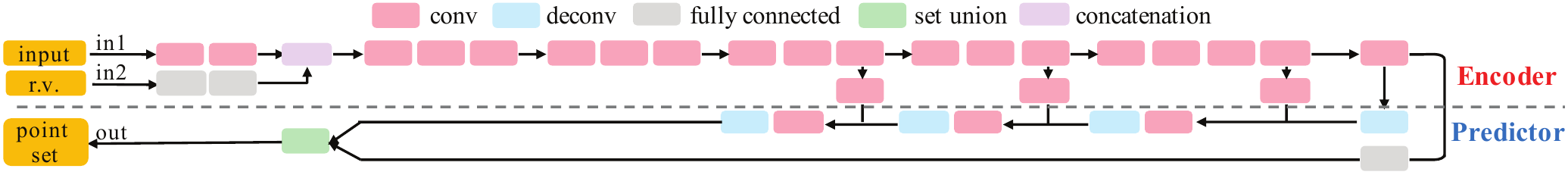}
    \put(40,-0.5){two prediction branch version}
  \end{overpic}    
  \qquad\\  
  \qquad\\  
  \begin{overpic}[width=\linewidth, trim={0cm, 0.47cm, 0cm, 0cm}, clip,unit=1mm]{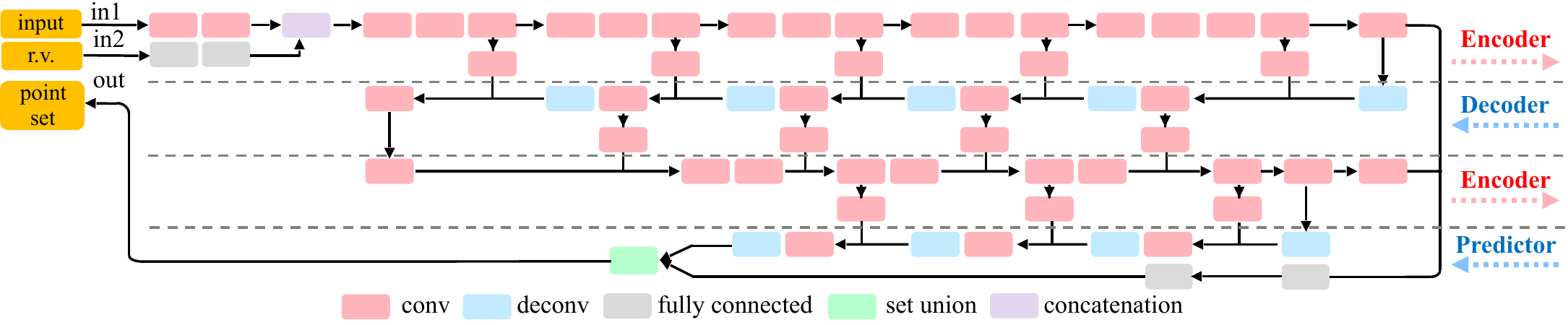}
    \put(47,-0.5){hourglass version}
  \end{overpic} 
  \caption{PointOutNet structure}
  \label{fig:pointnet}
\end{figure*}
\label{sec:problem}
Our goal is to reconstruct the \emph{complete} 3D shape of an object from a single 2D image (RGB or RGB-D). We represent the 3D shapes in the form of unordered point set $S=\set{(x_i, y_i, z_i)}_{i=1}^{N}$ where $N$ is a predefined constant. We observed that for most objects using $N=1024$ is sufficient to preserve the major structures. 

One advantage of point set comes from its unordered-ness. Unlike 2D based representations like the depth map no topological constraint is put on the represented object. Compared to 3D grids, the point set enjoys higher efficiency by encoding only the points on the surface. Also, the coordinate values $(x_i,y_i,z_i)$ go over simple linear transformations when the object is rotated or scaled, which is in contrast to the case in volumetric representations.

To model the problem's uncertainty, we define the groundtruth as a probability distribution $\prob(\cdot|\image)$ over the shapes conditioned on the input $\image$. 
In training we have access to one sample from $\prob(\cdot|\image)$ for each image $\image$.

We train a neural network $\network$ as a conditional sampler from $\prob(\cdot|\image)$: 
\vspace{-3mm}
\begin{align}
\label{eqn:main}
  \shape = \network(\image, r;\Theta)
\end{align}
where $\Theta$ denotes network parameter, $r\sim \mathbb{N}(\mathbf{0}, \mathbf{I})$ is a random variable to perturb the input~\footnote{Similar to the Conditional Generative Adversarial Network~\cite{mirza2014conditional}.}. During test time multiple samples of $r$ could be used to generate different predictions.

\section{Approach}
\subsection{Overview}

%
%

Our task of building a conditional generative network for point sets is challenging, due to the unordered form of representation and the inherent ambiguity of groundtruth. These challenges has pushed us to invent new architecture, loss function, and learning paradigm. Specifically, we have to address three subproblems: 

\para{Point set generator architecture}: Network to predict point set is barely studied in literature, leaving a huge open space for us to explore the design choices. Ideally, a network should make the best use of its data statistics and possess enough representation power. We propose a network with two prediction branches, one enjoys high flexibility in capturing complicated structures and the other exploits geometric continuity. Its representation power is further boosted by an hourglass structure. See Sec~\ref{sec:method:network}. 

\para{Loss function for point set comparison}: For our novel type of prediction, point set, it is unclear how to measure the distance between the prediction and groundtruth. We introduce two distance metrics for point sets -- the Chamfer distance and the Earth Mover's distance. We show that both metrics are differentiable almost everywhere and can be used as the loss function, but has different properties in capturing shape space. See Sec~\ref{sec:method:loss}.

\para{Modeling the uncertainty of groundtruth}: Our problem of 3D structural recovery from a single image is ill-posed, thus the ambiguity of groundtruth arises during the train and test time. It is fundamentally important to characterize the ambiguity of groundtruth for a given input, and practically desirable to be able to generate multiple predictions. Surprisingly, this goal can be achieved tactfully by simply using the $\min$ function as a wrapper to the above proposed loss, or by a conditional variational autoencoder.  See Sec~\ref{sec:method:gan}.




\subsection{Point Set Prediction Network}
\label{sec:method:network}
%
%
The task of building a network for point set prediction is new. We design a network with the goal of possessing strong representation power for complicated structures, and make the best use of the statistics of geometric data. 
To introduce our network progressively, we start from a simple version and gradually add components.


As in Fig~\ref{fig:pointnet} (top), our network has an encoder stage and a predictor stage. The encoder maps the input pair of an image $\image$ and a random vector $r$ into an embedding space. The predictor outputs a shape as an $N\times 3$ matrix ${\bf M}$, each row containing the coordinates of one point.

The encoder is a composition of convolution and ReLU layers; in addition, a random vector $r$ is subsumed so that it perturbs the prediction from the image $I$. We postpone the explanation of how $r$ is used to Sec~\ref{sec:method:gan}. 
The predictor generates the coordinates of $N$ points through a fully connected network. Though simple, this version works reasonably well in practice. 

We further improve the design of the predictor branch to better accommodate large and smooth surfaces which are common in natural objects. The fully connected predictor as above cannot make full use of such natural geometric statistics, since each point is predicted independently. The improved predictor in Fig~\ref{fig:pointnet} (middle) exploits this geometric smoothness property. 

This version has two parallel predictor branches -- a fully-connected (fc) branch and a deconvolution (deconv) branch. The fc branch predicts $N_1$ points as before. The deconv branch predicts a 3 channel image of size $H\times W$, of which the three values at each pixel are the coordinates of a point, giving another $H\times W$ points. Their predictions are later merged together to form the whole set of points in ${\bf M}$. Multiple skip links are added to boost information flow across encoder and predictor. 

With the fc branch, our model enjoys high flexibility, showing good performance at describing intricate structures. With the deconvolution branch, our model becomes not only more parameter parsimonious by weight sharing; but also more friendly to large smooth surfaces, due to the spatial continuity induced by deconv and conv. Refer to Sec~\ref{sec:exp:analysis} for experimental evidences. 

To pursue even better performance, we introduce the hourglass version in Fig~\ref{fig:pointnet} (bottom), inspired by \cite{newell2016stacked}. This deep network conducts the encoding-decoding operations recurrently, thus has stronger representation power and can mix global and local information better. 

Above introduces the design of our network $\mathbb{G}$ in Eq~\ref{eqn:main}. To train this network, however, we still need to design a proper loss function for point set prediction, and enable the role $r$ for multiple candidates prediction. We explain in the next two sections.
\subsection{Distance Metric between Point Sets}
\label{sec:method:loss}

A critical challenge is to design a good loss function for comparing the predicted point cloud and the groundtruth. To plug in a neural network, a suitable distance must satisfy at least three conditions: 1) differentiable with respect to point locations; 2) efficient to compute, as data will be forwarded and back-propagated for many times; 3) robust against small number of outlier points in the sets (e.g. Hausdorff distance would fail).

We seek for a distance $d$ between subsets in $\R^3$, so that the loss function $L(\{S^{pred}_i\}, \{S^{gt}_i\})$ takes the form
\begin{align}
    L(\{S^{pred}_i\}, \{S^{gt}_i\}) = \sum d(S^{pred}_i, S^{gt}_i),
    \label{eqn:loss}
\end{align}
where $i$ indexes training samples, $S_i^{pred}$ and $S_i^{gt}$ are the prediction and groundtruth of each sample, respectively.

We propose two candidates: Chamfer distance (CD) and Earth Mover's distance (EMD)~\cite{rubner2000earth}.

\paragraph{Chamfer distance} We define the Chamfer distance between $S_1, S_2\subseteq \R^3$ as:
\begin{align*}
d_{CD}(S_1, S_2)=\sum_{x\in S_1}\min_{y\in S_2} \|x-y\|^2_2+\sum_{y\in S_2}\min_{x\in S_1} \|x-y\|^2_2
\end{align*}
In the strict sense, $d_{CD}$ is not a distance function because triangle inequality does not hold. We nevertheless use the term ``distance'' to refer to any non-negative function defined on point set pairs. For each point, the algorithm of CD finds the nearest neighbor in the other set and sums the squared distances up.   Viewed as a function of point locations in $S_1$ and $S_2$, $\mbox{CD}$ is continuous and piecewise smooth. The range search for each point is independent, thus trivially parallelizable. Also, spatial data structures like KD-tree can be used to accelerate nearest neighbor search. Though simple, CD produces reasonable high quality results in practice.

\paragraph{Earth Mover's distance} 

Consider $S_1, S_2\subseteq \R^3$ of equal size $s=|S_1|=|S_2|$. The EMD between $A$ and $B$ is defined as:
\begin{align*}
d_{EMD}(S_1, S_2)=\min_{\phi:S_1\rightarrow S_2} \sum_{x\in S_1} \|x-\phi(x)\|_2
\end{align*}
where $\phi:S_1\rightarrow S_2$ is a bijection.

The EMD distance solves an optimization problem, namely, the assignment problem. For all but a zero-measure subset of point set pairs, the optimal bijection $\phi$ is unique and invariant under infinitesimal movement of the points. Thus EMD is differentiable almost everywhere. In practice, exact computation of EMD is too expensive for deep learning, even on graphics hardware. We therefore implement a $(1+\epsilon)$ approximation scheme given by \cite{bertsekas1985distributed}. We allocate fix amount of time for each instance and incrementally adjust allowable error ratio to ensure termination. For typical inputs, the algorithm gives highly accurate results (approximation error on the magnitude of $1\%$). The algorithm is easily parallelizable on GPU.

\paragraph{Shape space}
Despite remarkable expressive power embedded in the deep layers, neural networks inevitably encounter uncertainty in predicting the precise geometry of an object. Such uncertainty could arise from limited network capacity, insufficient use of input resolution, or the ambiguity of groundtruth due to information loss in 3D-2D projection. Facing the inherent inability to resolve the shape precisely, neural networks tend to predict a ``mean'' shape averaging out the space of uncertainty. The mean shape carries the characteristics of the distance itself.

\begin{figure}[t!]
\centering
\includegraphics[width=0.9\linewidth]{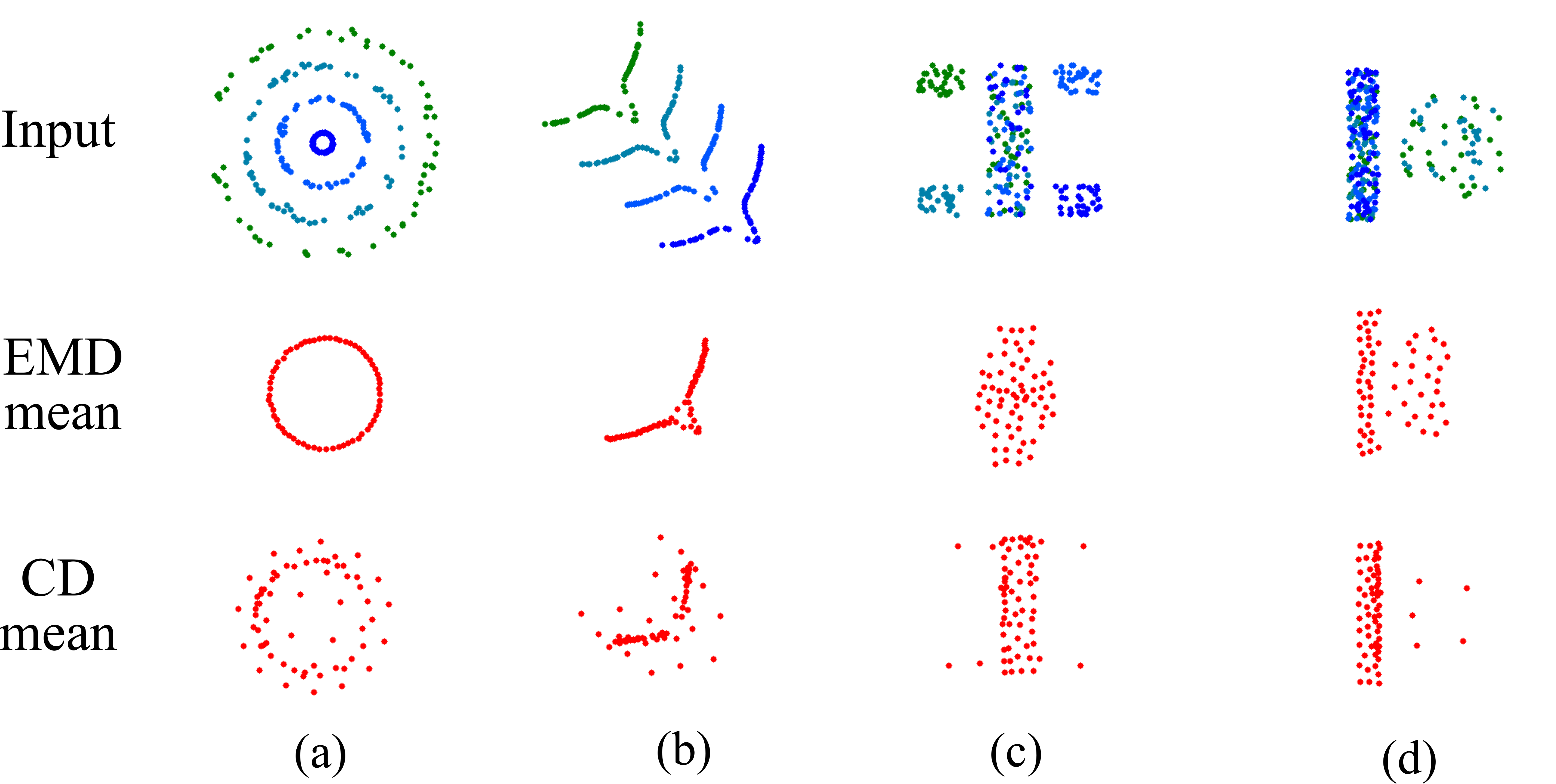}
\caption{Mean-shape behavior of EMD and CD. The shape distributions are (a) a circle with varying radius; (b) a spiky arc moving along the diagonal; (c) a rectangle bar, with a square-shaped attachment allocated randomly on one of the four corners; (d) a bar, with a circular disk appearing next to it with probability 0.5. The red dots plot the mean shape calculated according to EMD and CD accordingly.}
\label{fig:mean}
\end{figure}

In Figure~\ref{fig:mean}, we illustrate the distinct mean-shape behavior of EMD and CD on synthetic shape distributions, by minimizing
$\mathrm{E}_{s\sim \mathbb{S}}[L(x,s)]$
through stochastic gradient descent, where $\mathbb{S}$ is a given shape distribution, $L$ is one of the distance functions. 

In the first and the second case, there is a single continuously changing hidden variable, namely the radius of the circle in (a) and the location of the arc in (b). EMD roughly captures the shape corresponding to the mean value of the hidden variable. In contrast CD induces a splashy shape that blurs the shape's geometric structure. In the latter two cases, there are categorical hidden variables: which corner the square is located at (c) and whether there is a circle besides the bar (d). To address the uncertain presence of the varying part, the minimizer of CD distributes some points outside the main body at the correct locations; while the minimizer of EMD is considerably distorted.

\subsection{Generation of Multiple Plausible Shapes}
\label{sec:method:gan}
%
%

Our problem solves an ill-posed problem of 3D structural recovery from a single projection. Posed as a regression problem, ambiguity of the prediction arises at test time -- the depth for visible parts is under-determined, and the geometry for invisible parts has to be hallucinated by guessing. In a statistical view, reasonable predictions from the input image form a distribution.  Reflected in the training set, two images that look alike may have rather different groundtruth shapes. Recall the discussion in the previous section -- the ambiguity of groundtruth shape may significantly affect the trained predictor, as the loss function \eqref{eqn:loss} induces our model to predict the mean of possible shapes.  

\begin{figure}[t!]
  \centering
  \includegraphics[width=0.8\linewidth]{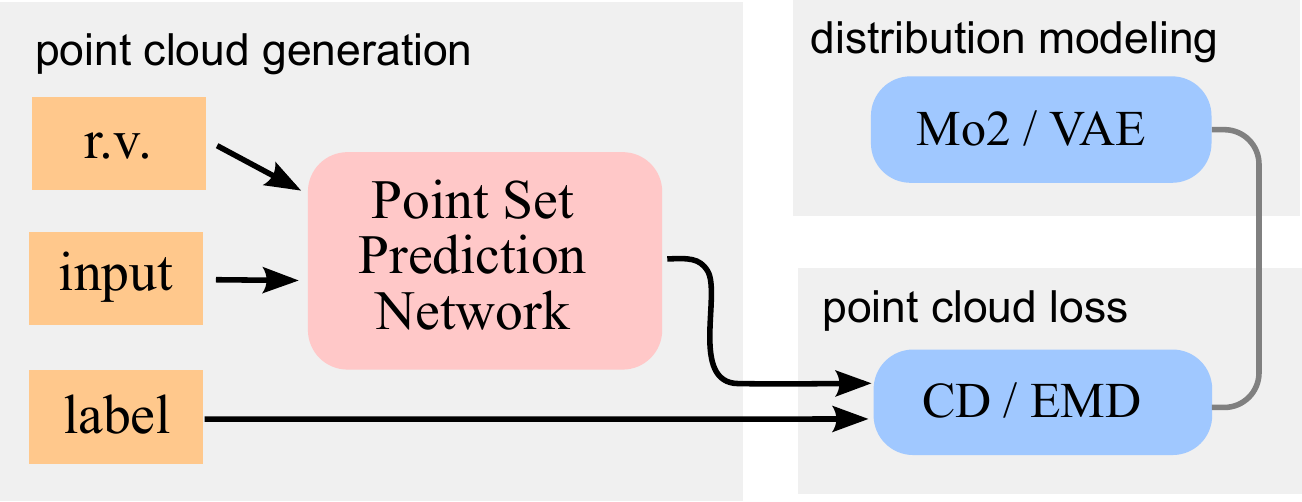}
  \caption{System structure. By plugging in distributional modeling module, our system is capable of generating multiple predictions. }
  \label{fig:network}
\end{figure}

To better model the uncertainty or inherent ambiguity (e.g. unseen parts in the single view), we enable the system to generate distributional output. We expect that the random variable $r$ passed to $\mathbb{G}$ (see Eq~\eqref{eqn:main}) would help it explore the groundtruth distribution, in analogy to conditional GAN (CGAN)~\cite{mirza2014conditional}. However, naively plugging $\mathbb{G}$ from Eq~\eqref{eqn:main} into Loss~\eqref{eqn:loss} to predict $S_i^{pred}$ won't work, as the loss minimization will nullify the randomness. It is also unclear how to make CGAN work in our scenario, as building a discriminator that directly consumes a point set is itself an open problem. 

The problem can be solved by more complex frameworks like VAE, where we can incorporate secondary input channels (e.g. another view). However, we find practically a simple and effective method for uncertainty modeling: the MoN loss. We train our network by minimizing a loss function as below:
\begin{equation}
    \begin{aligned}
    \underset{\Theta}{\mbox{minimize}} 
    && 
    \sum_k 
        \min_{
            \substack{r_j\sim \mathbb{N}(\mathbf{0}, \mathbf{I})\\1\le j\le n}
        }
        \{
            d(\mathbb{G}(I_k, r_j;\Theta), S_k^{gt})
        \}
    \end{aligned} 
    \label{eqn:gan}
\end{equation}    

We explain the rationale behind Problem~\eqref{eqn:gan} here. Given an image $I_k$, $\mathbb{G}$ makes $n$ predictions by perturbing the input with $n$ random vectors $r_j$. Intuitively, we expect that one of the predictions will be close to the groundtruth $S_k^{gt}$ given by the training data, meaning that the minimum of the $n$ distances between each prediction and the groundtruth must be small. 

We name this loss as Min-of-N loss (MoN), since it comes from the minimum of $n$ distances. Any of the point set regression networks in Fig~\ref{fig:pointnet} can be plugged into the meta network in Fig~\ref{fig:network} incorporating the MoN loss. In practice, we find that setting $n=2$ already enables our method to well explore the groundtruth space.  Please refer to Sec~\ref{sec:exp:gan} for experiment results.

\begin{figure}
\centering
\includegraphics[width=\linewidth]{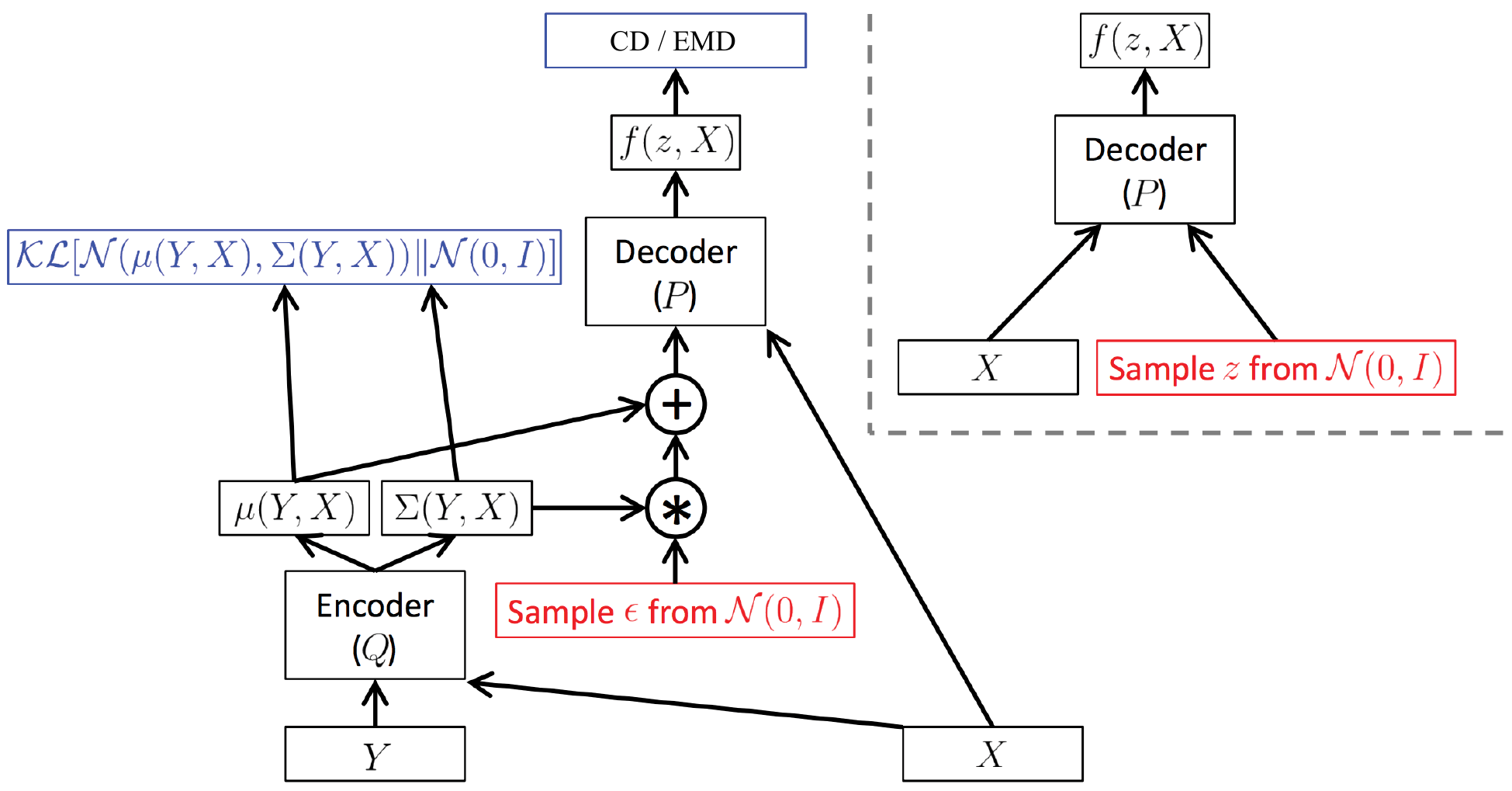}
\caption{Network for conditional variational autoencoder shape sampler $P(S|X)$. Left: a training-time conditional variational autoencoder implemented as a feedforward neural network. Here, $Y$ is the volumetric form of the groundtruth shape $S$, whereas $f(z, X)$ is the point cloud form of the predicted shape for $S$. Right: the same model at test time. (Modified from Doersch et al.~\cite{doersch2016tutorial})}
\label{fig:VAE}
\end{figure}

An alternative way to achieve the conditional shape sampler is by a conditional variational autoencoder. For more details about variational autoencoders, please refer to \cite{doersch2016tutorial}.  Fig~\ref{fig:VAE} shows the system architecture for training and testing a conditional variational autoencoder $P(S|X)$ in our case. Here, $X$ is the input image and $S$ is the \emph{point cloud} representation of the groundtruth 3D shape. At training time, each input image $X$ will be augmented by a random variable that is conditioned on $Y$, which takes the \emph{volumetric} representation of the groundtruth shape $S$. A 3D convolutional network is used as the encoder $Q$ (see \cite{maturana2015voxnet} for a good reference of 3D conv networks). Therefore, a local proximity in the embedding space contains the variations of possible groundtruth 3D shapes.

\section{Experiment}
\label{sec:exp}




\begin{figure}[t!]
  \centering
  \includegraphics[width=\linewidth]{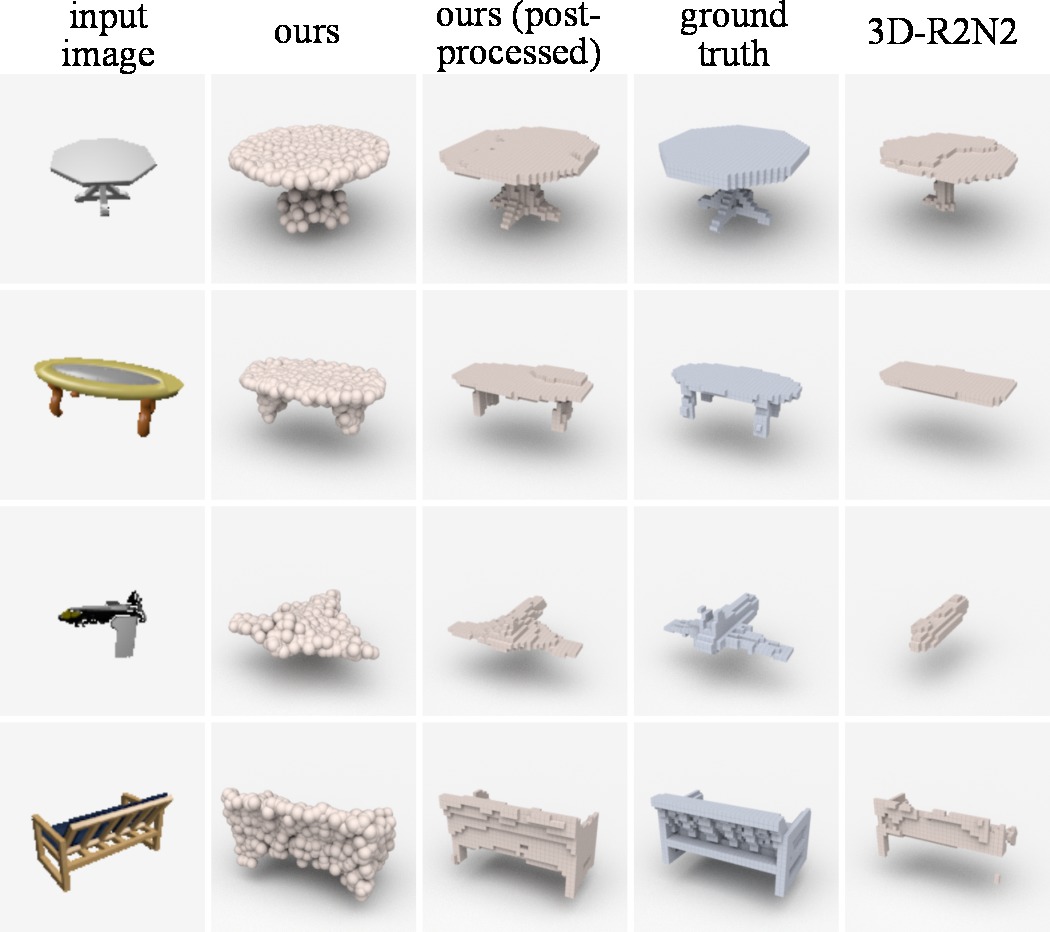}
  \caption{Visual comparison to 3D-R2N2. Our method better preserves thin structures of the objects. }\label{fig:visual_comparison}
\end{figure}

\subsection{Training Data Generation by Synthesis}
\label{sec:exp:traindata}
To start, we introduce our training data preparation. We take the approach of rendering 2D views from CAD object models. 
Our models are from the ShapeNet dataset~\cite{shapenet2015}, containing large volume of manually cleaned 3D object models with textures. Concretely we used a subset of 220K models covering 2,000 object categories. The use of synthesized data has been adopted in a number of existing works~\cite{choy20163d,rezende2016unsupervised}.

\begin{figure}[t!]
  \centering
  \includegraphics[width=0.9\linewidth]{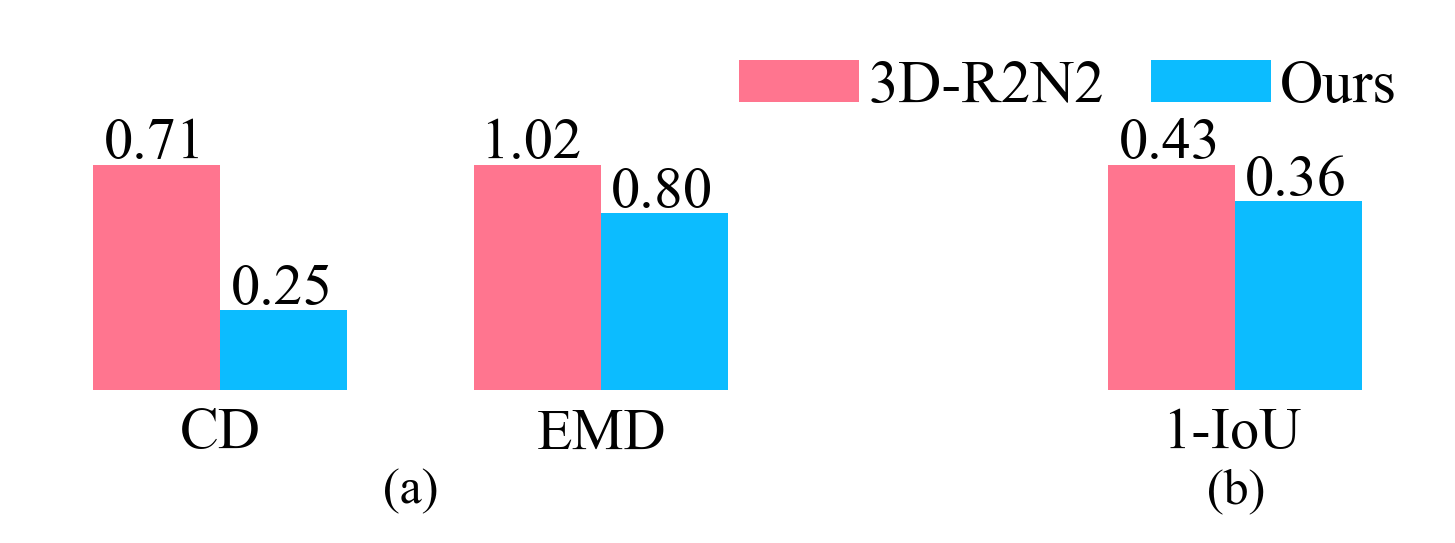}
  \caption{Quantitative comparison to 3D-R2N2. (a) Point-set based metrics CD and EMD. (b) Volumetric representation based metric 1 - IoU. Lower bars indicate smaller errors. Our method gives better results on all three metrics. }\label{fig:comparison}
\end{figure}
For each model, we normalized the radius of its bounding hemi-sphere to unit $1$ and aligned their ground plane. Then each model was rendered into 2D images according to the Blinn-Phong shading formula with randomly chosen environmental maps. In our experiments we used a simple local lightening model for the sake of computation time. However, it is straight-forward to extend our method to incorporate global illumination algorithms and more complex backgrounds. 


\subsection{3D Shape Reconstruction from RGB Images}

\label{sec:exp:rgb}
\paragraph{Comparison to state-of-the-art}
We compare our work to 3D-R2N2\cite{choy20163d} which is the state-of-the-art in deep learning based 3D object generation. 3D-R2N2 reconstructs 3D from single or multi-view images into a volumetric representation. To enable the comparison we re-trained our networks on the dataset used by 3D-R2N2's authors. The results are compared under three different metrics CD, EMD and IoU (intersection over union). 
In 3D-R2N2 only IoU values are reported, so we used the trained network provided by the authors to compute their predictions. To compute CD and EMD, their predicted and ground truth volumes are sampled by iterative farthest point sampling~\cite{eldar1997farthest} to a discrete set of points with the same cardinality as ours. We post-processed our point-set into a volumetric one with the same resolution as in 3D-R2N2 when computing IoU. Refer to Sec~\ref{sec:impl_details} for details.

In Fig~\ref{fig:comparison} we report the result of our network compared with the single-view 3D-R2N2. To determine the absolute scale of CD and EMD we define unit $1$ as $1/10$ of the length of the 3D grid used to encode the ground truth shape in 3D-R2N2's dataset. Though not directly trained by IoU, our network gives significantly better performance under all three measures. 

\begin{table}[t!]
\centering
{
  \begin{tabular}{l|c|c|c|c}
  \hline
  \multirow{2}{*}{category} & Ours & \multicolumn{3}{|c}{3D-R2N2} \\  
  \cline{2-5}
   & 1 view & 1 view & 3 views & 5 views
  \\
  \hline
  \hline
  plane & {\textbf{0.601}} & 0.513 & 0.549 & 0.561 \\
%
  bench & {\textbf{0.550}} & 0.421 & 0.502 & 0.527 \\
  cabinet & {0.771} & 0.716 & 0.763 & \textbf{0.772}  \\
  car & {0.831} & 0.798 & 0.829 & \textbf{0.836} \\
  chair & {0.544} & 0.466 & 0.533 & \textbf{0.550}  \\
  monitor & {0.552} & 0.468 & 0.545 & \textbf{0.565} \\
  lamp & {\textbf{0.462}} & 0.381 & 0.415 & 0.421 \\
  speaker & {\textbf{0.737}} & 0.662 & 0.708 & 0.717 \\
  firearm & {\textbf{0.604}} & 0.544 & 0.593 & 0.600 \\
  couch & {\textbf{0.708}} & 0.628 & 0.690 & 0.706 \\
  table & {\textbf{0.606}} & 0.513 & 0.564 & 0.580 \\
  cellphone & {0.749} & 0.661 & 0.732 & \textbf{0.754} \\
  watercraft & {\textbf{0.611}} & 0.513 & 0.596 & 0.610 \\
  \hline
  mean & {\textbf{0.640}} & 0.560 & 0.617 & 0.631 \\
  \hline
  \end{tabular}
  }
  \caption{3D reconstruction comparison (per category). Notice that in the single view reconstruction setting we achieved higher IoU in all categories. The mean is taken category-wise. For 8 out of 13 categories, our results are even better than 3D-R2N2 given 5 views.}\label{tab:compare_category}
\end{table}

We report the IoU value for each category as in~\cite{choy20163d}. From Table~\ref{tab:compare_category},
we can see that the for single view reconstruction the proposed method consistently achieves higher IoU in all categories. 3R-R2N2 is also able to predict 3D shapes from more than one views. On many categories our method even outperforms the 3D-R2N2's prediction given 5 views.

To further contrast the two methods, we visualize some typical examples. As stated in~\cite{choy20163d}, their method often misses thin features of objects (e.g. legs of furnitures). We surmise that this is due to their volumetric representation and voxel-wise loss function which unduly punishes mispositioned thin structures. In contrast, our point-cloud based objective function encourages the preservation of fine structures and makes our predictions more structurally plausible.

\subsection{3D Shape Completion from RGBD Images}
\label{sec:exp:depth}
\begin{figure}[th!]
  \centering
  \includegraphics[width=0.9\linewidth]{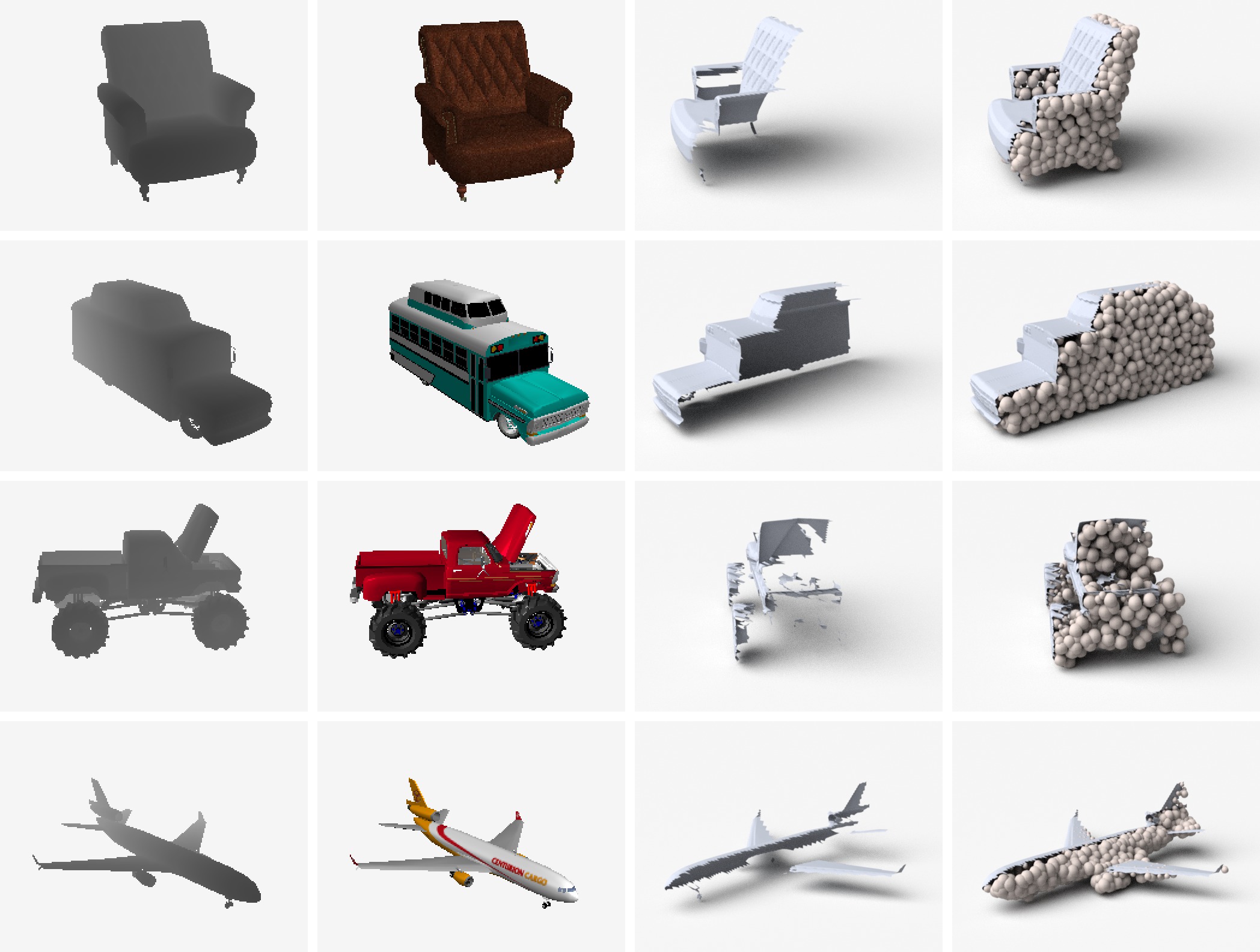}
  \caption{Shape completion from a single RGBD image.}\label{fig:shape_completion}
\end{figure}
One interesting feature of our approach is that we can easily inject additional input information into the system. When the neural network is given RGBD input our system can be viewed as a 3D shape completion method. Fig~\ref{fig:shape_completion} visualizes examples of the predictions.

The neural network successfully guesses the missing parts of the model. By using the shape priors embedded in the object repository, the system can leverage cues of both symmetry (e.g. airplanes should have symetric sides) and functionality (tractors should have wheels). The flexible representation of point set facilitates the resolution of the object's general shape and topology. More fine-grained methods that directly exploit local geometric cues could be cascaded after our predictions to enrich higher frequency details.

\subsection{Predicting Multiple Plausible Shapes}
\label{sec:exp:gan}
\begin{figure}[t!]
  \centering
  \includegraphics[width=0.9\linewidth]{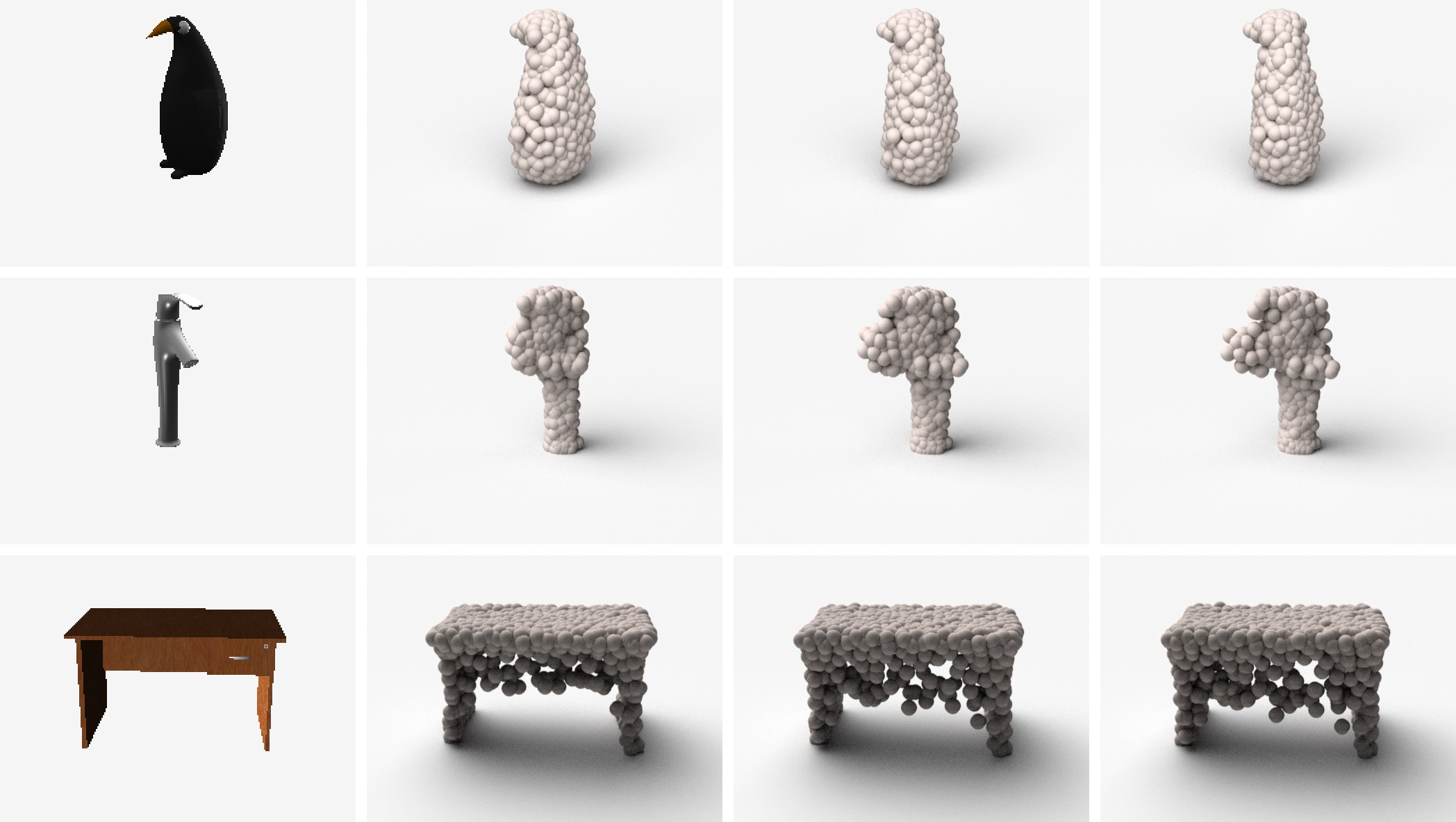}
  \caption{Multiple predictions for a single input image. The point sets are visualized from different view points (top row: half side view, middle row: side view, bottom row: back view) to better reveal the difference.}\label{fig:deformation}
\end{figure}

\begin{figure}
\centering
\includegraphics[width=0.9\linewidth]{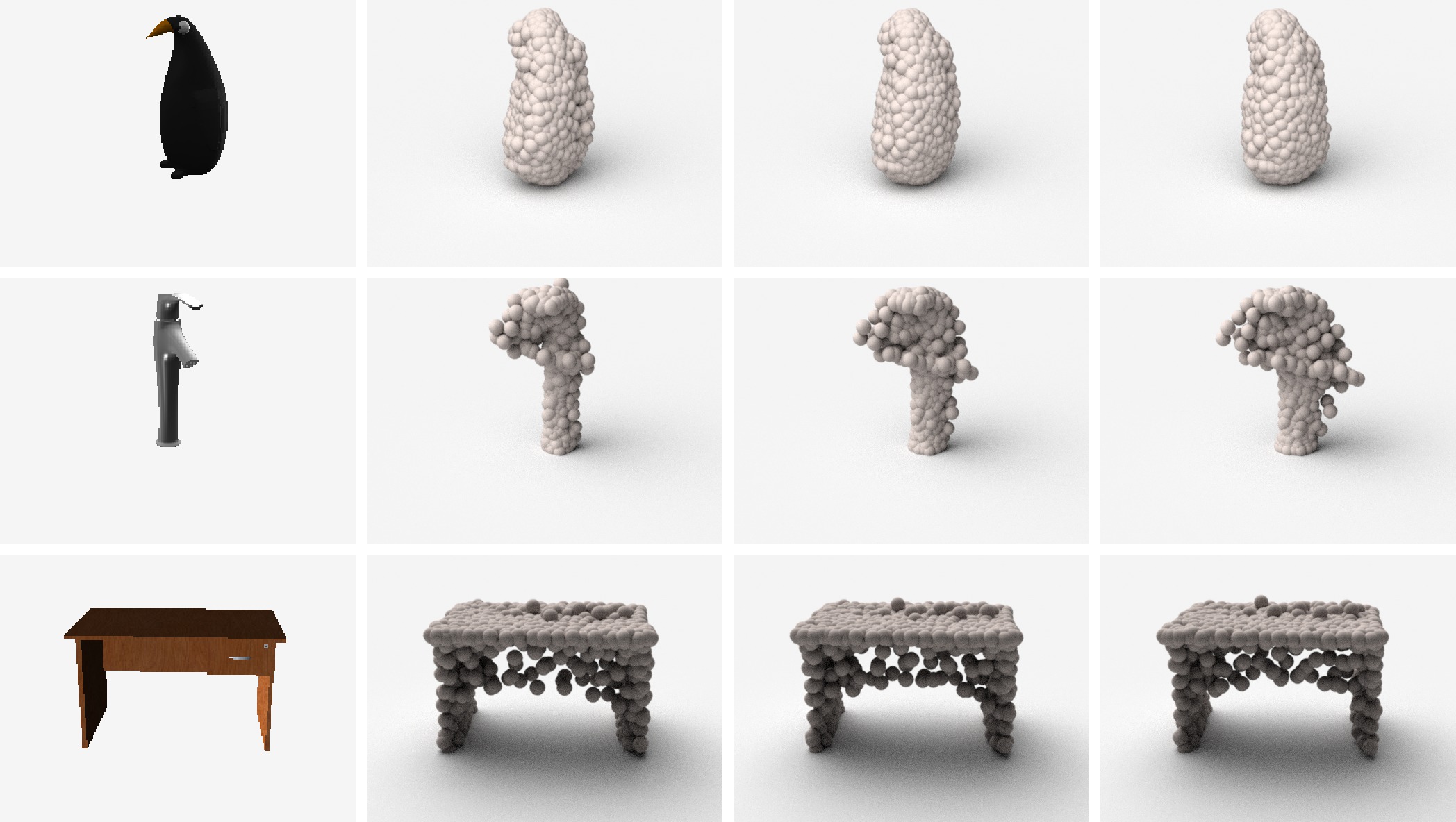}
\caption{Result obtained by VAE training. Top: half-side view; middle: side view; bottom: back view.}
\label{fig:show_vae}
\end{figure}

The randomness in our network enables prediction of different shapes given the same input image. To show this, we take the RGB image as the input. During training we handle randomness by using either the Mo2 or the VAE method. At test time when the ground truth is unknown, the random numbers are sampled from the predefined distribution.

Fig~\ref{fig:deformation} plots examples of the set of predictions of our method. The network is able to reveal its uncertainty about the shape or the ambiguity in the input. Points that the neural network is certain about its position moves little between different predictions. Along the direction of ambiguity (e.g. the thickness of the penguin's body) the variation is significantly larger. In this figure we trained our network with Mo2 and Chamfer Distance. In Fig~\ref{fig:show_vae}, we visualize the results of VAE. Compared to the result of Mo2, the prediction of VAE looks plumper; however, it also captures the local directions of ambiguity in the shape.

\begin{figure}[h!]
  \centering
  \includegraphics[width=0.9\linewidth]{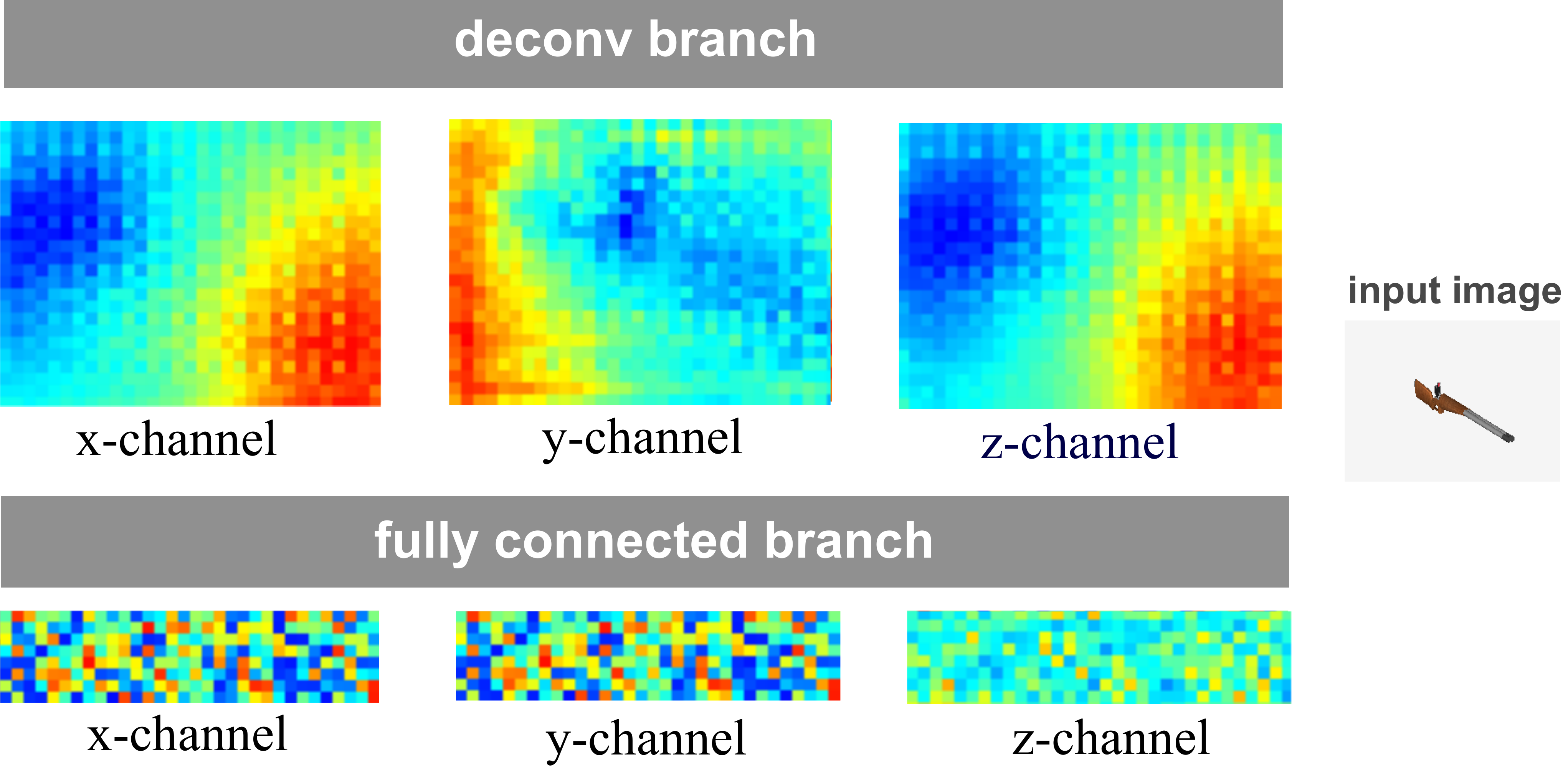}
  \caption{Visualization of the channels.}\label{fig:vis_deconv_channels}
  \vspace{-1em}
\end{figure}
\subsection{Network Design Analysis}

\begin{figure}[!]
  \centering
  \includegraphics[width=0.9\linewidth]{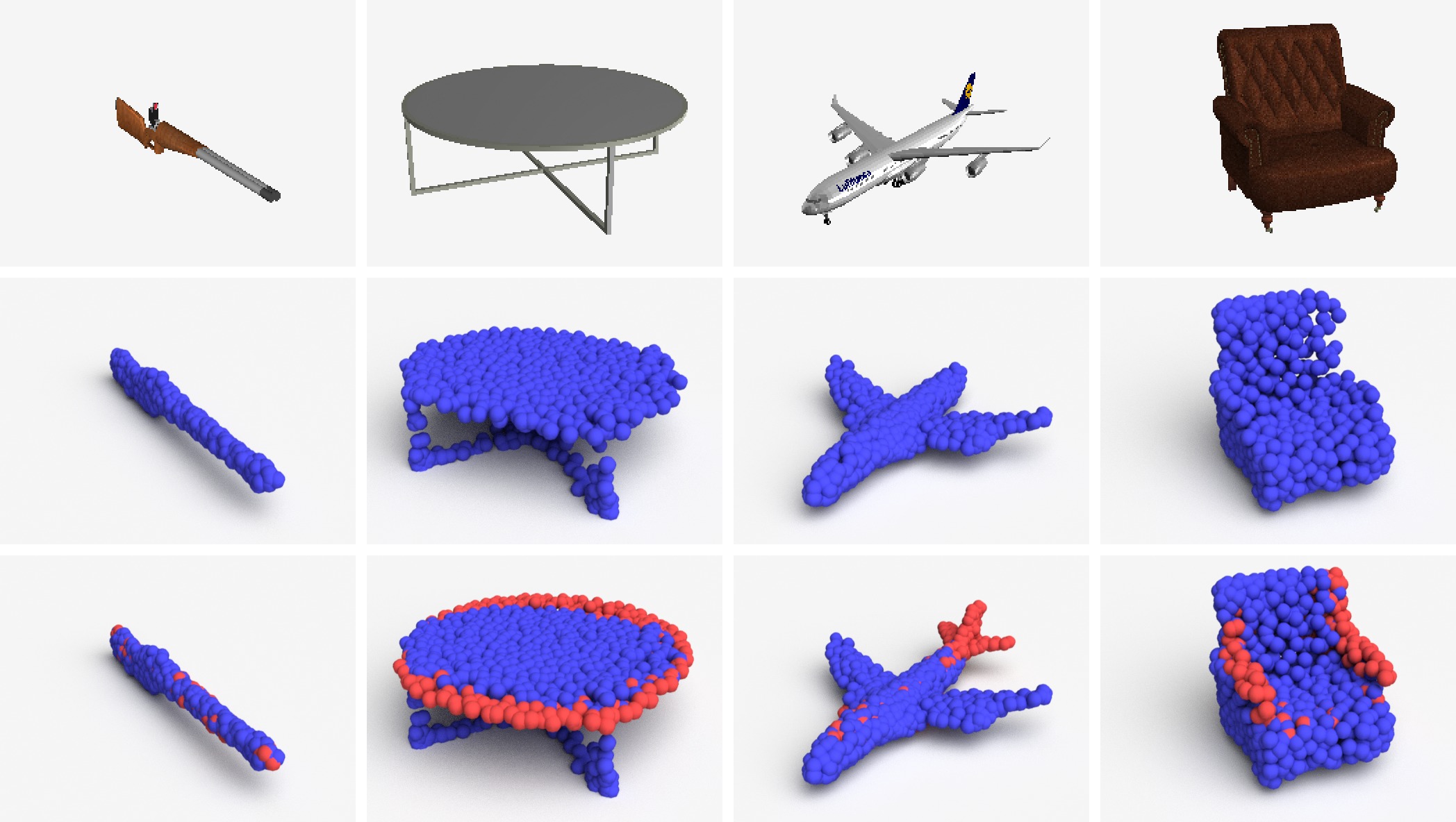}
  \caption{Visualization of points predicted by the deconvolution branch (blue) versus the fully connected branch (red).}\label{fig:vis_deconv_vs_fc}
  \vspace{-1em}
\end{figure}
\label{sec:exp:analysis}
\paragraph{Effect of combining deconv and fc branches for reconstruction}

\begin{figure*}[t!]
   \centering
   \vspace{1em}
   \includegraphics[width=0.9\linewidth]{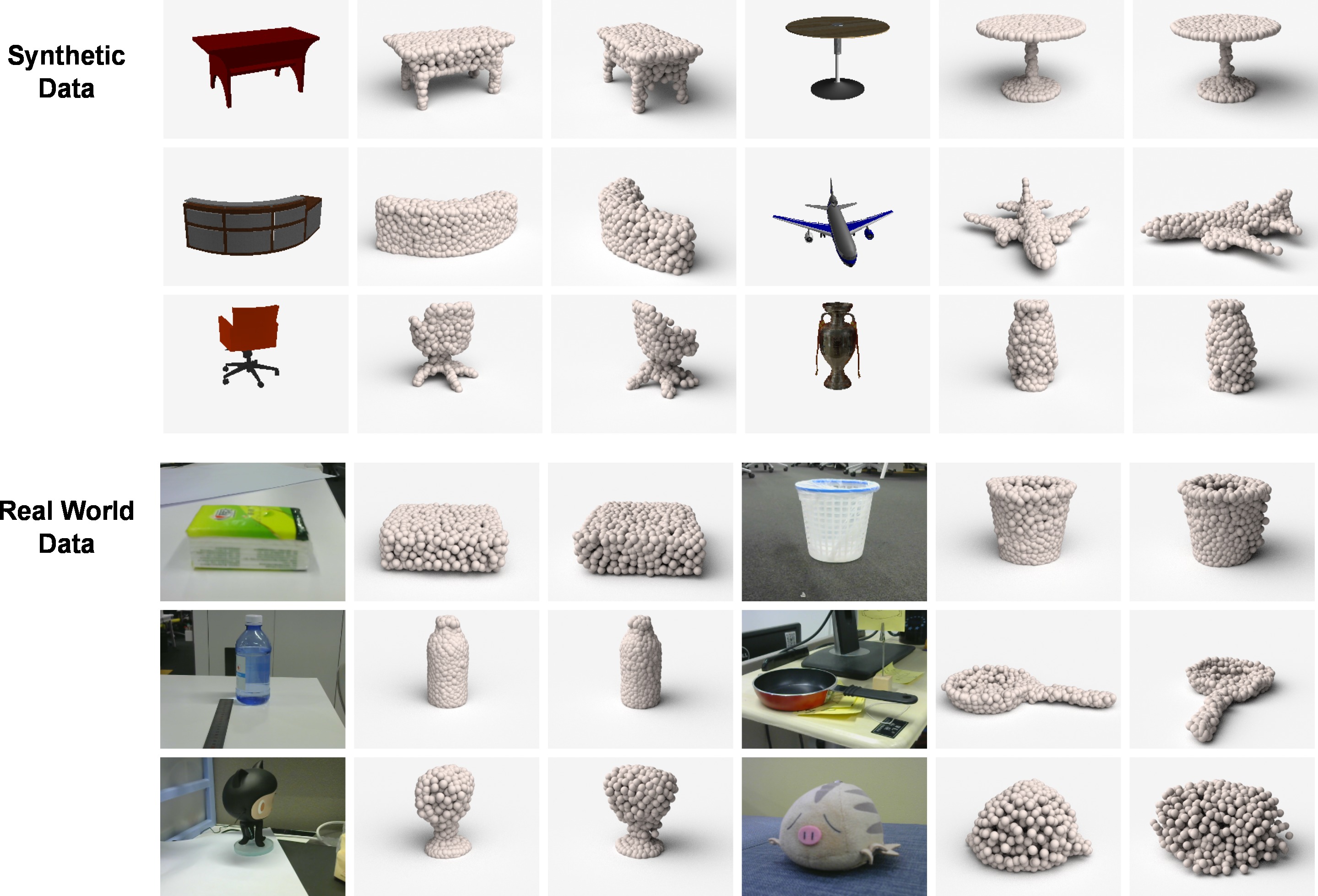}
   \caption{Visualization of predictions on synthetic and real world data.}\label{fig:more_examples}
 \end{figure*}
\begin{figure}[t!]
  \centering
  \includegraphics[width=\linewidth]{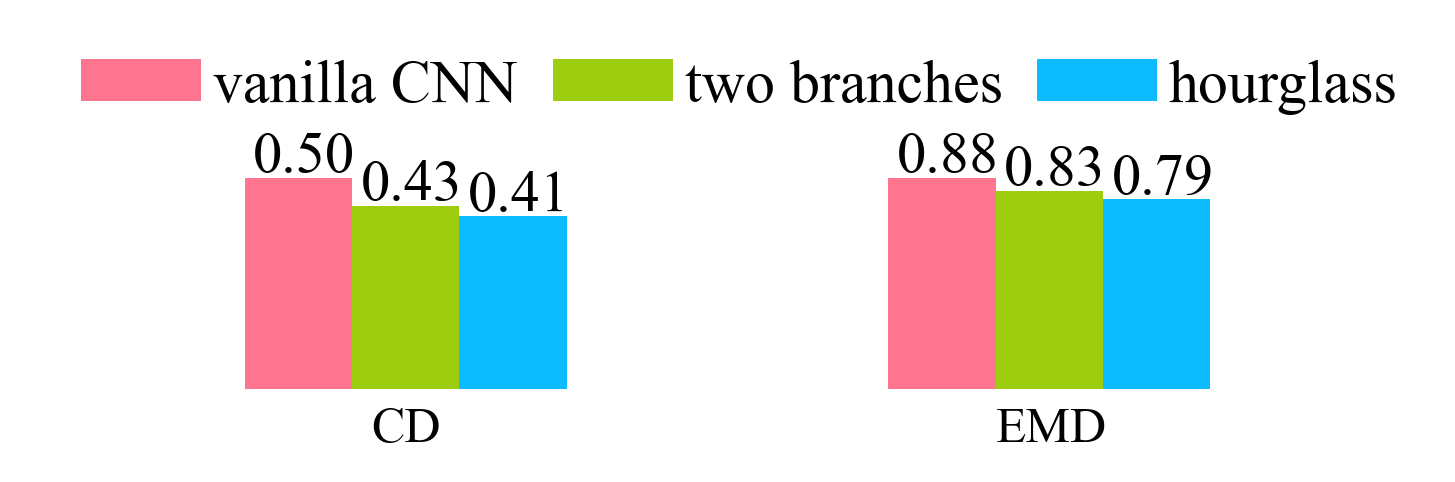}
  \caption{Comparison of different networks by Chamfer Distance (CD) and Earth Mover Distance (EMD). More complex network gives slightly better results. }\label{fig:compare_networks}
  \vspace{-1em}
\end{figure}

We compared different designs of the neural network architectures. The performance values are reported based on our own rendered training set.
As shown in Fig~\ref{fig:compare_networks}, the introduction of deconvolution significantly improves performance. Stacking another hourglass level also gives performance gain.

We further visualize the output of the deconv branch and fully connected branch separately to gain a better understanding of their functions. In Fig~\ref{fig:vis_deconv_channels} the values in the x, y and z channels are plotted as 2D images for one of the models. In the deconv branch the network learns to use the convolution structure to constructs a 2D surface that warps around the object. In the fully connected branch the output is less organized as the channels are not ordered.

In Fig~\ref{fig:vis_deconv_vs_fc} we render the two set of predictions in 3D space. The deconv branch is in general good at capturing the ``main body'' of the object, while the fully connected branch complements the shape with more detailed components (e.g. tip of gun, tail of plane, arms of a sofa). This reveals the complementarity of the two branches. The predefined weights sharing and node connectivity endow the deconv branch with higher efficiency when they are congruent with the desired output's structure. The fully connected branch is more flexible but the independent control of each point consumes more network capacity.


\paragraph{Analysis of distance metrics}

\begin{figure}[ht!]
  \centering
  \includegraphics[width=\linewidth]{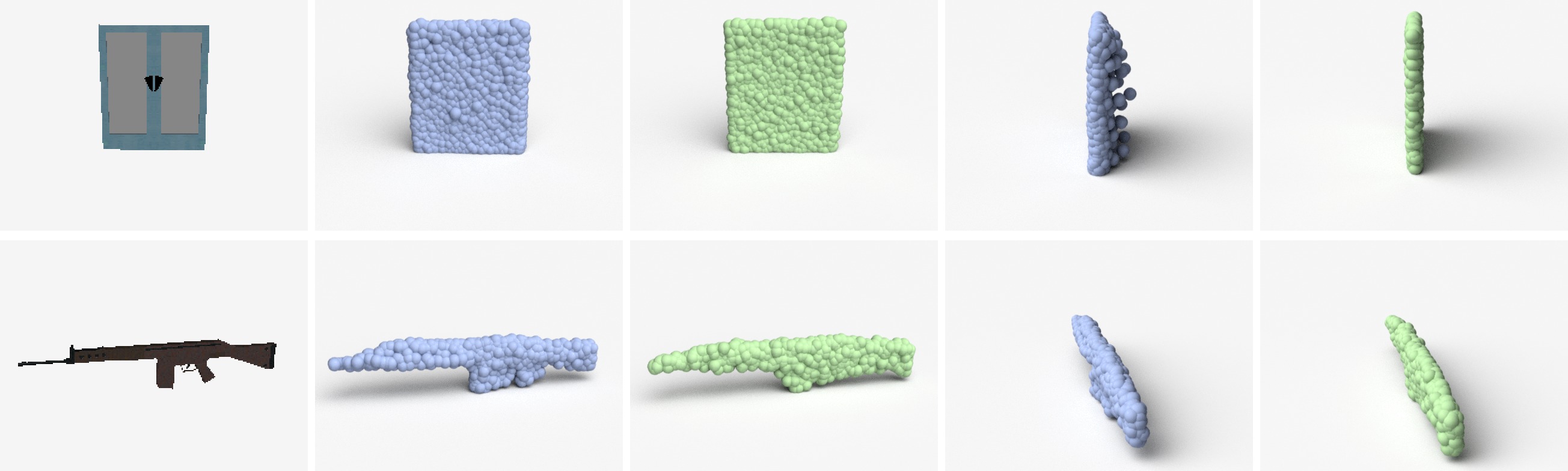}
  \caption{Comparison of predictions of networks trained by CD (blue, on the left) and EMD (green, on the right). }\label{fig:emd_vs_chamfer}
  \vspace{-1em}
\end{figure}

Different choices of the loss functions have distinct effect on the network's prediction pattern. Fig~\ref{fig:emd_vs_chamfer} exemplifies the difference between two networks trained by CD and EMD correspondingly. The network trained by CD tends to scatter a few points in its uncertain area (e.g. behind the door) but is able to better preserve the detailed shape of the grip. In contrast, the network trained by EMD produces more compact results but sometimes overly shrinks local structures. This is in line with experiment on synthetic data.

\subsection{More results and application to real world data}
\label{sec:exp:more}
 
\begin{figure*}
	\centering
	\includegraphics[width=\linewidth]{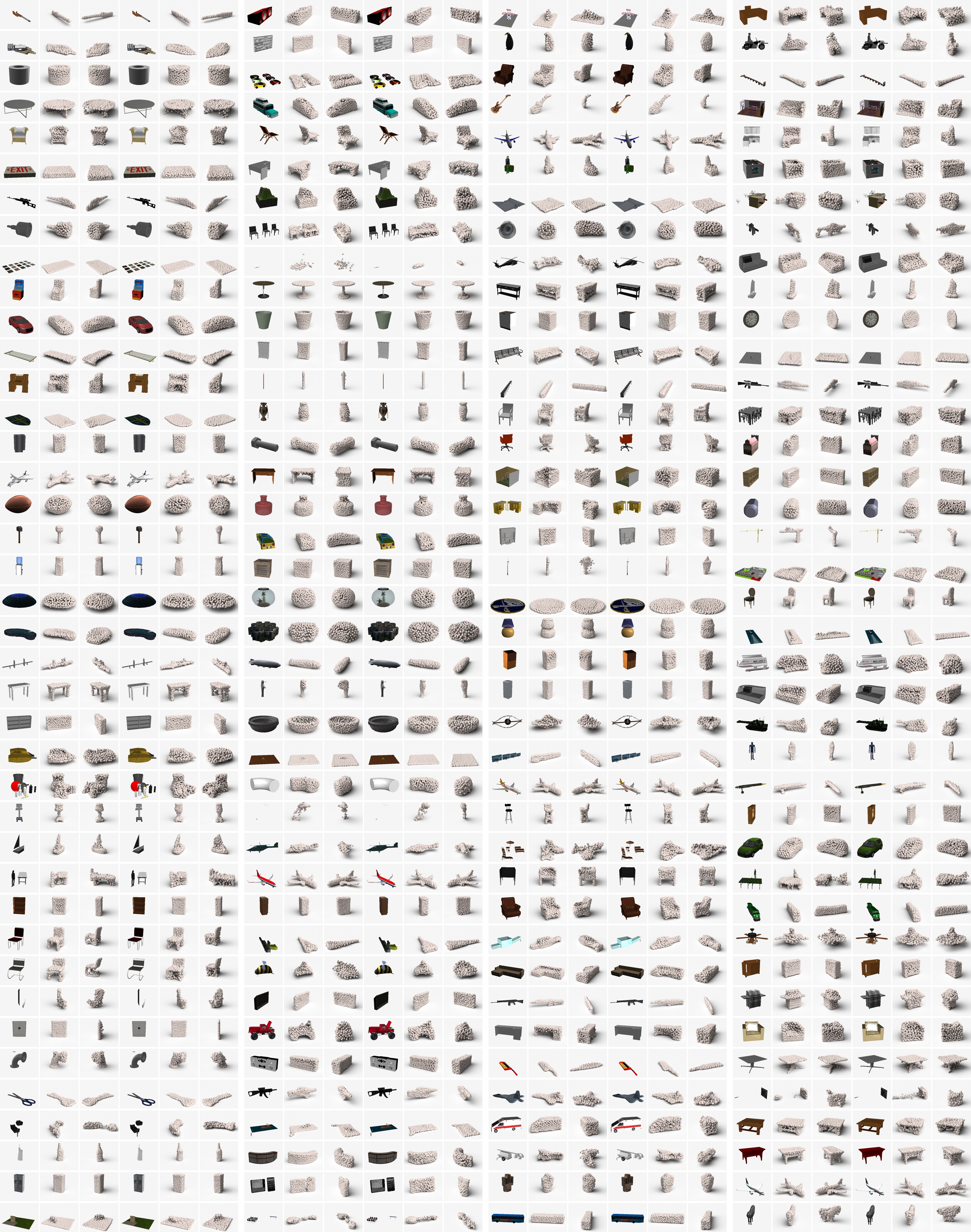}
	\caption{First 5 mini-batches of our validation set. Result obtained by CD is on the left, EMD on the right.}
	\label{fig:show_all}
\end{figure*}

Fig~\ref{fig:more_examples} lists more example predictions on both synthetic data and real world photos. For real world photo, we mask out background pixels to indicate the object. Our algorithm gives promising result though trained on synthetic data only.
We plot the reconstruction results of the first 5 mini-batches (160 cases in total) of our validation set at the end of this paper in Fig~\ref{fig:show_all}. Results produced by the network trained by CD and EMD are compared side-by-side. Owing to the diversity in the ShapeNet dataset, our system is able to handle a variety of object types.

\subsection{Analysis of human ability for single view 3D reconstruction}
\begin{figure}
\centering
\includegraphics[width=1.0\linewidth]{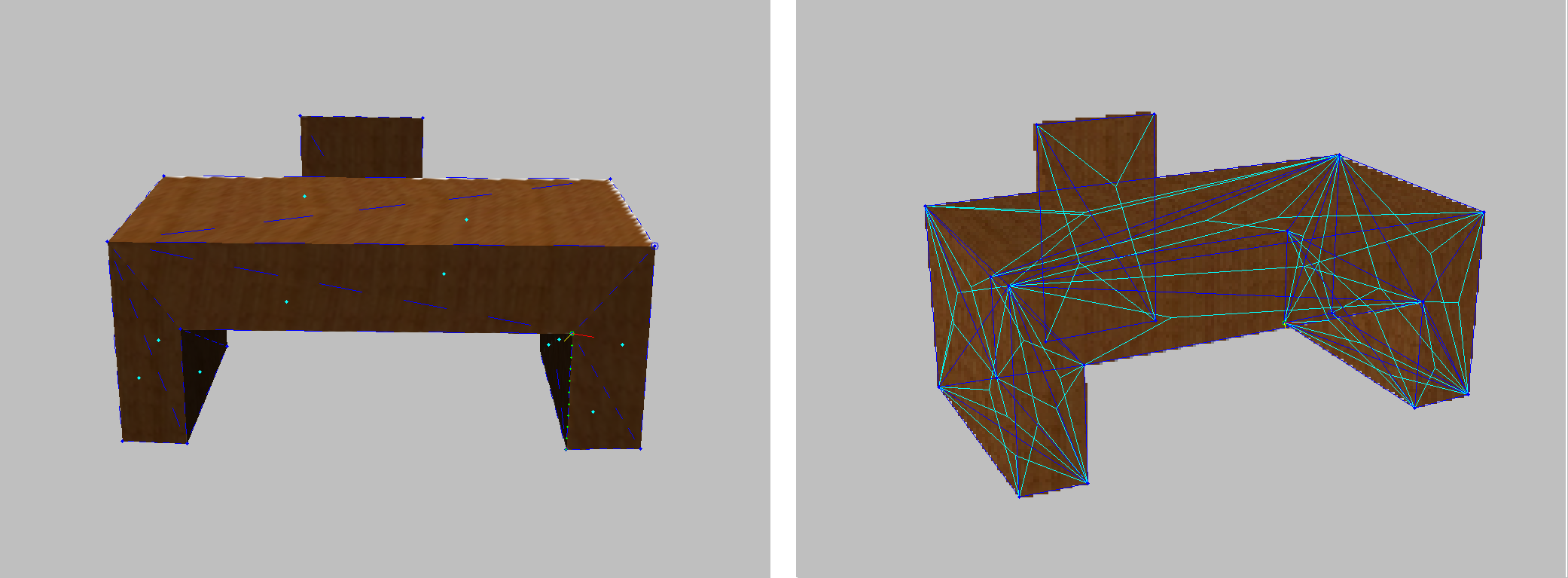}
\caption{GUI tool used to manually model the objects. The user can change view point, edit vertex positions and connectivity in the 3D view (bottom). We also overlaid a wire-frame rendering of the object on the input image (top) to facilitate alignment.}
\label{fig:gui_tool}
\end{figure}

We conducted human study to provide reference to our current CD and EMD values reported on the rendered dataset. We provided the human subject with a GUI tool to create a triangular mesh from the image. The tool (see Fig~\ref{fig:gui_tool}) enables the user to edit the mesh in 3D and to align the modeled object back to the input image. In total 16 models are created from the input images of our validation set. $N=1024$ points are sampled from each model.

\begin{figure}
\centering
\includegraphics[width=1.0\linewidth]{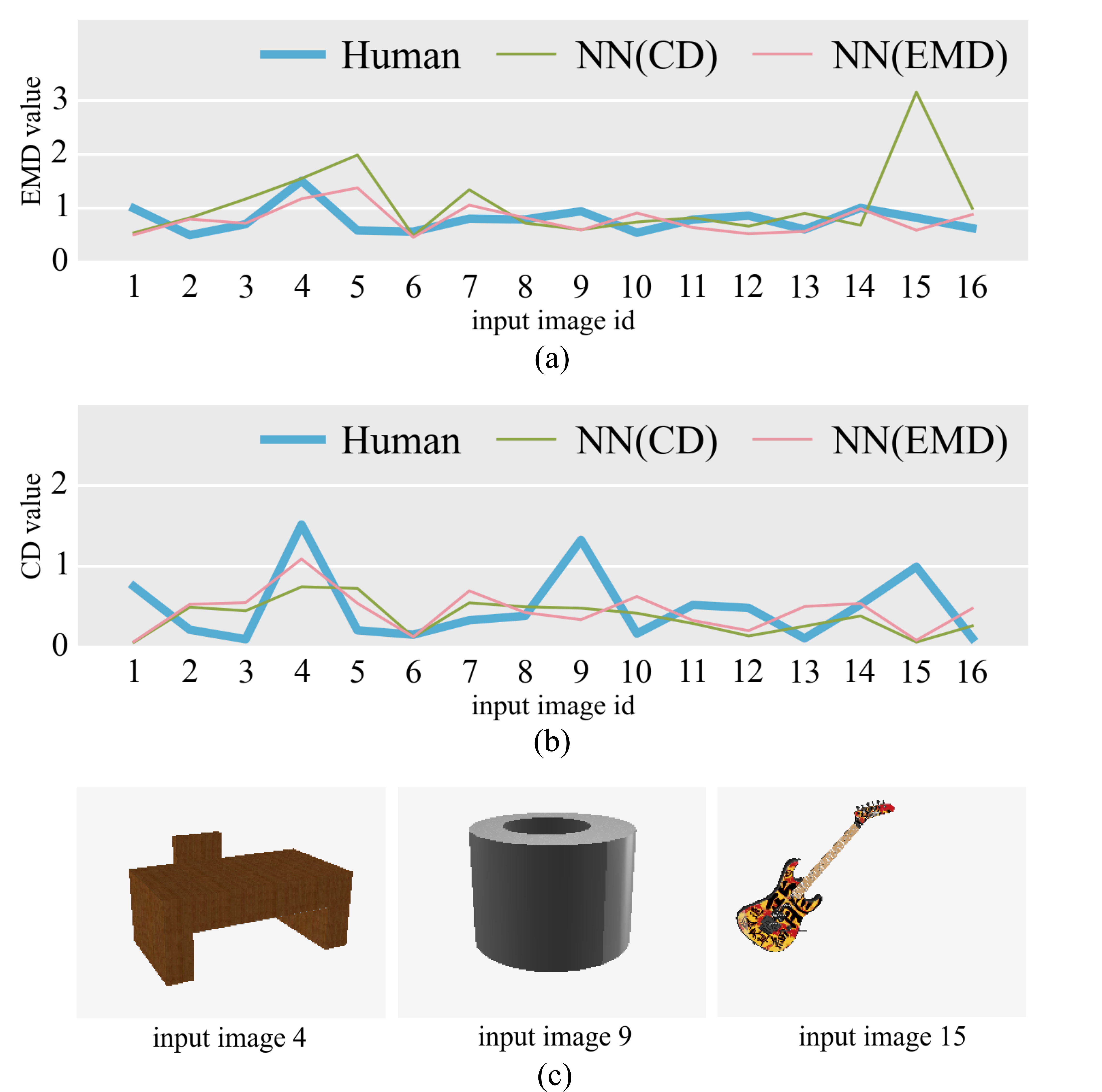}
\caption{Comparison of reconstructions generated by the human subject, the neural network trained with CD and the neural netword trained with EMD on 16 input images in the validation set. (a) Comparison of EMD value. (b) Comparison of CD value. (c) Input images numbered 4, 9 and 19 on which the human subject performs poorly.}
\label{fig:human_number}
\end{figure}

As shown in Fig~\ref{fig:human_number}, both the EMD and the CD values of the network's reconstruction are on par with human's manual creation for most of the cases. We observed that the human subject mainly used cues of gravity direction (legs of chairs should touch the ground) and symmetry to infer the object's shape. As illustrated in input image number 4, 9 and 15, when the object is partially occluded (the table blocks the chair), ambiguous (it is unclear whether the can has a bottom) or manifests inadequate geometric cues (the guitar has non-polygonal shape and does not sit on the ground) the human subject performs poorly. The neural network trained by EMD performs reasonably well under both metrics. However, because CD emphasises only on the best matching point, the network trained by CD does not always produce predictions of uniform density and suffers high EMD value in some cases.

\subsection{Analysis of failure cases}

\begin{figure}
\centering
\includegraphics[width=\linewidth]{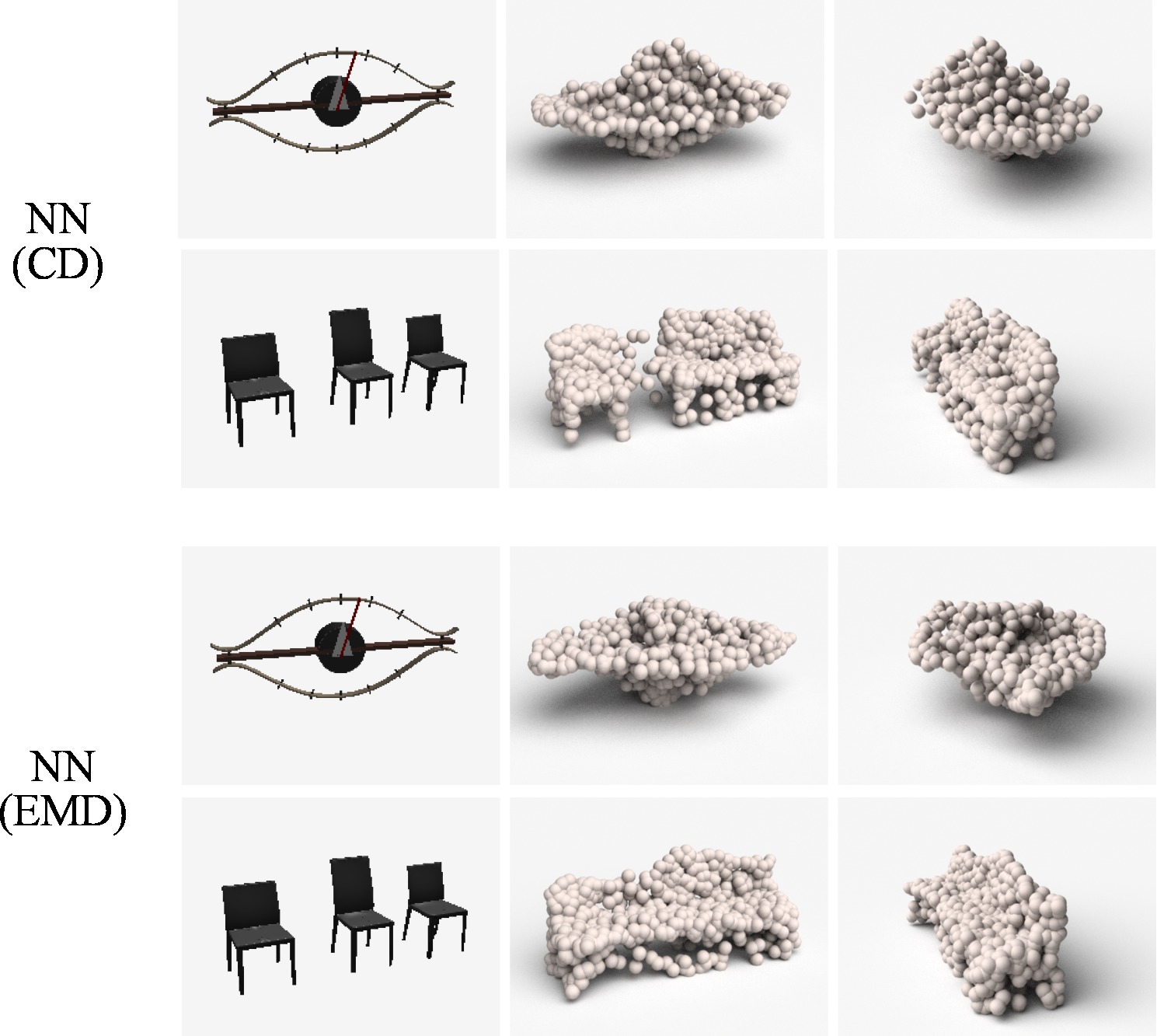}
\caption{Examples of failure cases of our method on the validation set. Top: results of the neural network trained by CD. Bottom: results of the neural network trained by EMD. Both networks give unsatisfactory results.}
\label{fig:failure_case}
\end{figure}

We visualize representative failure cases of our method on our rendered validation set. There are two trends, each exemplified by one input case in Fig~\ref{fig:failure_case}. In the first kind of failure cases, the neural network is presented with a shape that it has completely no idea about. Then the networks tried to explain the input by something similar (a plane without wings?) but fundamentally wrong. In the second kind of failure cases, the neural network sees a composition of multiple objects. Because we have not implemented any detection or attention mechanism, the networks produce distorted output.

\subsection{Implementation details}
\label{sec:impl_details}
\paragraph{Network parameter and training}
Our network works on input images of 192x256. The deconv branch produces 768 points, which correspond to a 32x24 three-channel image. The fully connected branch produces 256 points. The convolutional layer has 16 feature maps in the highest resolution, and the number of channels are doubled after each decrease in resolution. We use strided convolution instead of max-pooling to increase speed. The training program is implemented in TensorFlow. 300000 gradient steps are taken, each computed from a minibatch of 32. Adam is used as the optimizer. We observed that the training procedure is smooth even without batch normalization. All activation functions are relu.

\paragraph{Post processing}
We use a local method to post process the point cloud into a volumetric representation. First, the point cloud is registered into the 32x32x32 grid with bilinear interpolation. This can be think of as interpreting the points as 1x1x1 cubes and averaging the intersection volume with each grid cell (the occupancy representation). Then each voxel exams a local neighborhood to determine the final value. We implement this as a trained 3D convolutinoal neural network with 6 layers of 3x3x3 convolutions. This post-processing network is trained by IoU on the same training partition as the point cloud generation network. In order to compensate for difference in point density among objects of different volumes, we trained another network to predict the object's volume. The predicted volume is concatenated with the registered occupancy as the 3D conv network's input. Using the point cloud generation network trained by either EMD or CD to is enough to outperform 3D-R2N2's result. The maximum performance as reported in the main paper is obtained by feeding both     network's prediction into the post processing network. We also notice that the volume prediction network is not necessary to outperform 3D-R2N2. However, it consistently gives performance gain, so we kept this component in our experiments.

\section{Discussion}
Though presented as an application paper, we have touched two fundamental problems: First, how to generate an orderless set of entities. Towards building generative models for more sophisticated combinatorial data structures such as graphs, knowing how to generate a set may be a good starting point. Second, how to capture the ambiguity of the groundtruth in a regression problem . Other than 3D reconstruction, many regression problems may have such inherent ambiguity. Our construction of the MoN loss by wrapping existing loss functions may be generalizable to these problems.

{\bibliographystyle{ieee}
\bibliography{si2pc}
}

\end{document}